\def\year{2021}\relax
\documentclass[letterpaper]{article} % DO NOT CHANGE THIS
\usepackage{aaai21}  % DO NOT CHANGE THIS
\usepackage{times}  % DO NOT CHANGE THIS
\usepackage{helvet} % DO NOT CHANGE THIS
\usepackage{courier}  % DO NOT CHANGE THIS
\usepackage[hyphens]{url}  % DO NOT CHANGE THIS
\usepackage{graphicx} % DO NOT CHANGE THIS
\usepackage{amsmath}
\usepackage{amssymb}
\usepackage{multirow}
\usepackage{multicol}
\usepackage{url}
\usepackage{graphicx}
\usepackage{algorithm}  
\usepackage{algorithmicx} 
\usepackage{algpseudocode}
\usepackage{natbib}  % DO NOT CHANGE THIS AND DO NOT ADD ANY OPTIONS TO IT
\usepackage{caption} % DO NOT CHANGE THIS AND DO NOT ADD ANY OPTIONS TO IT
\title{Semi-supervised Medical Image Segmentation through Dual-task Consistency}
\author{Xiangde Luo\textsuperscript{\rm 1,2}, 
Jieneng Chen\textsuperscript{\rm 3}, Tao Song\textsuperscript{\rm2}, Yinan Chen\textsuperscript{\rm2}, Guotai Wang\textsuperscript{\rm 1\thanks{Corresponding author}}, Shaoting Zhang\textsuperscript{\rm 1,2}
}
\affiliations{\\
\textsuperscript{\rm 1}University of Electronic Science and Technology of China, Chengdu, China\\
\textsuperscript{\rm 2} SenseTime Research, Shanghai, China \\
\textsuperscript{\rm 3}Tongji University, Shanghai, China\\
xiangde.luo@std.uestc.edu.cn, chenjn@tongji.edu.cn, songtao@sensetime.com, guotai.wang@uestc.edu.cn\\}

\begin{document}
\maketitle
\begin{abstract}
Deep learning-based semi-supervised learning (SSL) algorithms have led to promising results in medical images segmentation and can alleviate doctors' expensive  annotations by leveraging unlabeled data. However, most of the existing SSL algorithms in the literature tend to  regularize  the  model training by perturbing networks and/or data. Observing that multi/dual-task learning attends to various levels of information which have inherent prediction perturbation, we ask the question in this work: can we explicitly build task-level regularization rather than implicitly constructing networks- and/or data-level perturbation and then regularization for SSL? To answer this question, we propose a novel dual-task-consistency semi-supervised framework for the first time. Concretely, we use a dual-task deep network that jointly predicts a pixel-wise segmentation map and a geometry-aware level set representation of the target. The level set representation is converted to an approximated segmentation map through a differentiable task transform layer. Simultaneously, we introduce a dual-task consistency regularization between the level set-derived segmentation maps and directly predicted segmentation maps for both labeled and unlabeled data. Extensive experiments on two public datasets show that our method can largely improve the performance by incorporating the unlabeled data. Meanwhile, our framework outperforms the state-of-the-art semi-supervised learning methods. Code is available at: \url{https://github.com/HiLab-git/DTC}
\end{abstract}

\section{Introduction}
\par Accurate and robust segmentation of organs or lesions from medical images plays an essential role in many clinical applications such as diagnosis and treatment planning~\cite{masood2015survey}. With a large amount of labeled data, deep learning has achieved the state-of-the-art performance on automatic image analysis~\cite{long2015fully,chen2018encoder,song2020cpm}. For medical image, however, annotations are often expensive to acquire as both expertise and time are needed to produce accurate annotations, especially in 3D volumetric images. To reduce the labeling cost, recently, many methods are proposed to develop a high-performance model for medical image segmentation with less labeled data. For example, combining user interaction with deep neural network to perform image segmentation interactively can reduce the labeling efforts~\cite{bifseg,deepigeos}. Self-supervised learning approaches utilize unlabeled data to train models in a supervised manner to learn fundamental knowledge for knowledge transfer~\cite{zhu2020rubik}. Semi-supervised learning framework obtains high-quality segmentation results by learning from a limited amount of labeled data and a large set of unlabeled data directly~\cite{li2020transformation,qiao2018deep,zhou2019semi,xia20203d,masood2019automatic}. Weakly supervised learning methods learn from bounding boxes, scribbles or image-level tags for image segmentation rather than using pixel-wise annotation, which reduces the burden of annotation~\cite{dai2015boxsup, lin2016scribblesup,lee2019ficklenet}. In this work, we focus on the semi-supervised segmentation methods, as it is more practical to acquire a small set of fully annotated images with a large set of unannotated images.
% the performance of weakly supervised learning and self-supervised learning are much limited for segmentation of 3D medical images, and 

Many recent successful SSL methods ~\cite{yu2019uncertainty, li2020transformation,nie2018asdnet, li2020shape} incorporate unlabeled data by performing unsupervised consistency regularization. To be specific, they can either add small perturbations to the unlabeled samples and enforce the consistency between the model predictions on the original data and the perturbed data ~\cite{yu2019uncertainty, li2020transformation}, or just directly enforce the similar prediction distributions on the entire unlabeled dataset with an adversarial regularization ~\cite{nie2018asdnet, li2020shape}. Thus, we have learned that the essence of the discussed SSL works is to enforce the consistency on predictions of unlabeled data via a regularization term in loss function.

Among the aforementioned SSL works, it is delighted to see \citet{li2020shape} developed a multi-task network containing the pixel-wise and the shape-aware prediction branches, similar to previous fully supervised works \cite{wang2020deep, Xue2020ShapeAwareOS}. And for SSL, they consider only the shape branch to build the consistent constraints via an adversarial regularization to make prediction distributions on the entire unlabeled dataset be smooth, which still belongs to data-level regularization. We observe that various levels of information from different task branches can complement each other during training, while different focuses can lead to inherent prediction perturbation. For example, if  the predictions from pixel-wise branch and shape-aware branch are finally evaluated under the same criterion, we will definitely obtain different results i.e. the prediction perturbations between different tasks. Then we  ask  the most significant  question  in  this  work: can  we  explicitly build task-level regularization totally different from previous data-level regularization? Apparently the answer is yes, on the condition that the output of different task branches should be mapped/transformed to the same predefined space, where we are capable to explicitly enforce the consistency regularization between two prediction maps.

To this end, we propose a novel dual-task-consistency model for semi-supervised medical image segmentation. Our main idea is to build the consistency between a global-level level set function regression task and a pixel-wise classification task to take geometric constraints into account and utilize the unlabeled data. Our framework consists of three parts: the first part is dual-task segmentation network. Specifically, we model  a segmentation problem as two different representations (tasks): predicting a pixel-wise classification map and obtaining a global-level level set function where the zero level let is the segmentation contour. We use a two-branch network to predict these two representations, and using a CNN to predict level set function is inspired by~\cite{ma2020distance,ma2020TMI-LGAC,Xue2020ShapeAwareOS} to embed global information and geometric constraints into a network for better performance. The second part of this framework is a differentiable task transform layer. We use a smooth Heaviside layer~\cite{Xue2020ShapeAwareOS} to convert the level set function to a segmentation probability map in a differentiable way. The third part is a combination loss function for supervised and unsupervised learning, where we design a dual-task-consistency loss function to minimize the difference between the predicted pixels-wise segmentation probability map and the probability map converted  from the level set function, which can be used to boost the performance of fully supervised learning and also can be used to utilize the unlabeled data for unsupervised learning efficiently. Our proposed framework has been applied to two different semi-supervised medical image segmentation tasks: left atrium segmentation from MRI and pancreas segmentation from CT. Experimental results indicate that our proposed algorithm can improve segmentation accuracy, compared to other state-of-the-art semi-supervised segmentation methods. Overall, we present a simple yet efficient semi-supervised medical image segmentation method with dual-task consistancy, which leverages the unlabeled data by encouraging consistent predictions of the same input under different tasks. Our findings during experiments include:
\begin{itemize}

    \item[1)] In the fully supervised setting, our dual-task consistency regularization outperforms the separate and joint supervision of dual tasks.
     
    \item[2)] In the semi-supervised setting, the proposed framework outperforms state-of-the-art semi-supervised medical image segmentation frameworks on several clinical datasets. 
    
    \item[3)] Compared with existing methods, the proposed framework requires less training time and computational cost. Meanwhile, it is directly applicable to any semi-supervised medical image segmentation scene and can easily be extended to use additional tasks given that there exists a differentiable transform between/among tasks.
\end{itemize}

\section{Related Works}

\subsubsection{Semi-Supervised Medical Image Segmentation:}
For semi-supervised medical image segmentation, traditional methods mainly use hand-crafted features to design a model to perform segmentation, which includes the prior-based models~\citep{you2011segmentation} and the clustering-based models~\citep{portela2014semi}. The performance of the hand-crafted features-based models often relies on the hand-crafted features' representation capacity. For example, the prior-based models need to design the specific prior information for different organs, which can hardly generalize to other organs. The clustering-based models are often parameter-sensitive and not robust enough, which leads to the poor prediction for objects with large shape variance. 

With the ability to learn high-level semantic features automatically, deep learning has been widely used for medical image segmentation~\citep{ronneberger2015u}. Recently, almost all semi-supervised medical image segmentation frameworks are based on deep learning. \citet{bai2017semi} developed an iterative framework where in each iteration,  pseudo labels for unannotated images are predicted by the network and refined by a Conditional Random Field (CRF)~\citep{krahenbuhl2011efficient}, then the new pseudo labels are used to update the network. Using adversarial learning to utilize the unlabeled data is also a popular way for semi-supervised medical image segmentation. \citet{zhang2017deep} proposed a new deep adversarial network
(DAN) model for biomedical image segmentation by encouraging the segmentation of unannotated images to be similar to those of the annotated ones. \citet{yu2019uncertainty} extended the mean teacher model~\citep{tarvainen2017mean} with uncertainty map guidance for semi-supervised left atrium segmentation. \citet{li2020shape} introduced a shape-aware semi-supervised segmentation strategy to leverage the unlabeled data and to enforce a geometric shape constraint on the segmentation output. Differently, our method takes advantage of geometric constraints and dual-task-consistency, which is simple yet effective for semi-supervised medical image segmentation.

\begin{figure*}
    \centering
    \includegraphics[width=1.0\textwidth]{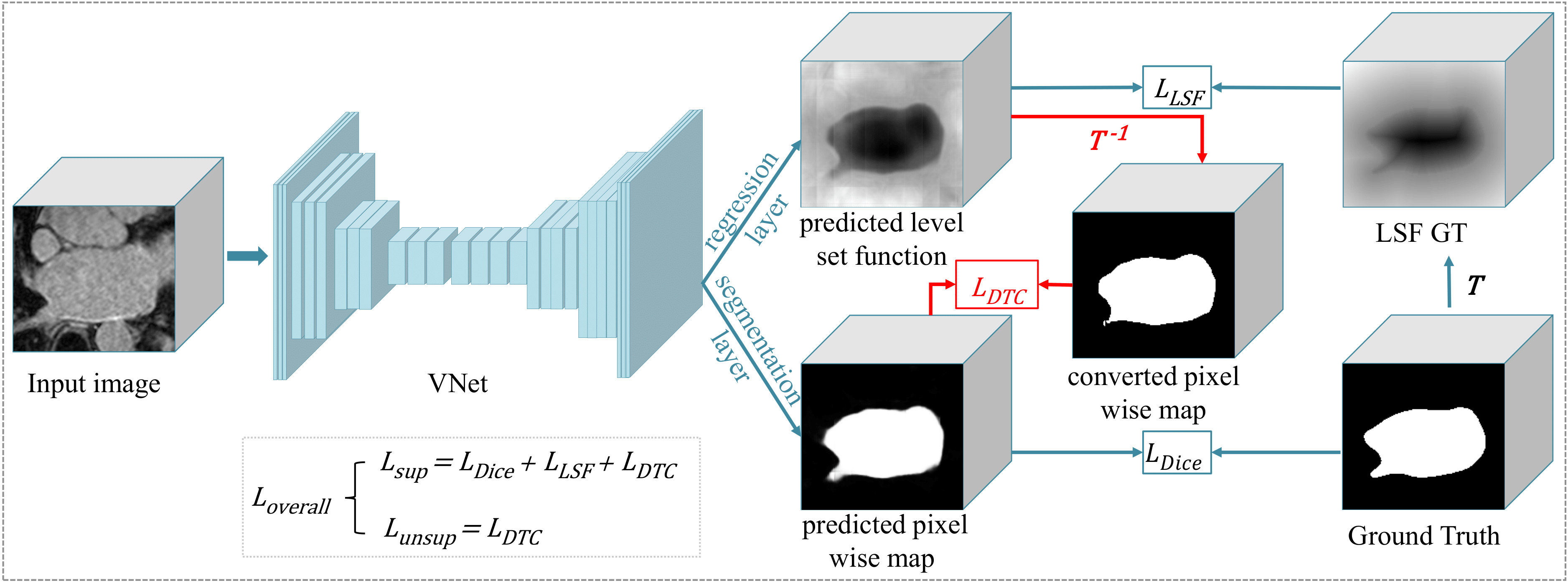}
    \caption{Overview of the proposed dual-task-consistency framework for semi-supervised medical image segmentation. The network consists of a pixel-wise classification head (task1) and a level set function regression head (task2), which employs a widely-used encoder-decoder network as the backbone, i.e., VNet~\cite{milletari2016v}. The model is optimized by minimizing supervised losses $L_{Dice}$, $L_{LSF}$ on labeled data and the dual-task-consistency loss $L_{DTC}$ on both unlabeled data and labeled data. The $T$ function is used to transform the ground truth label map into a level set representation for supervised training. The $T^{-1}$ function converts the level set function to a probability map to calculate the $L_{DTC}$.}
    \label{fig:framework}
\end{figure*}

\subsubsection{Consistency Regularization:}The consistency regularization plays a vital role in computer vision and image processing, especially in semi-supervised learning. For examples, \citet{sajjadi2016regularization} proposed a regularization with stochastic transformations and perturbations for deep semi-supervised learning, and learned from unlabeled images by minimizing the difference between the predictions of multiple passes of a training sample. \citet{tarvainen2017mean} introduced a teacher-student consistency model to make full use of the unlabeled data, where the student model learns from the teacher model by minimizing the segmentation loss on the labeled data and the consistency loss with respect to the targets from the teacher model on all input data. \citet{jeong2019consistency} used consistency constraints as a tool for enhancing detection performance by making full use of available unlabeled data. \citet{li2020transformation} introduced a transformation-consistent based semi-supervised segmentation method, which encourages consistent predictions of the network-in-training for the same input under different perturbations. However, these works just consider the consistency when the input under different perturbations and transformation, which ignore the consistency  of different tasks. In addition, these methods need to perform forward pass two or more times for calculating the consistency loss, which increases the computational cost and running time. 
% 需要重新修改下！！
More recently, \citet{zamir2020robust} utilized the consistency cross different tasks based on inference-path invariance, indicating it is promising to investigate task consistency. The limitation is that they require labeled data in a fully supervised manner and only studied on low-level vision tasks. 
% low-level vision task ? more easily transform?
In contrast to aforementioned methods, our framework aims to utilize the unlabeled data by minimizing the consistency between two tasks of a network, which considers the difference of different tasks and just needs to perform inference once. To the best of our knowledge, our work is the first to construct the task-consistency constraint for semi-supervised learning.

\section{Methods}
In this section, we introduce our proposed semi-supervised medical image framework based on dual-task-consistency.
The overall framework is illustrated in Figure~\ref{fig:framework}, which consists of two heads, the classification head for pixel-wise probability map and the regression head for level set representation of the target. The segmentation network takes a 3D medical image
as input, and predicts the level set function and pixel-wise probability map at the same time. As a segmentation result can be represented by both a pixel-level label map and a high-level contour related to a level set function, these two predictions should be consistent for the segmentation task. To utilize the unlabeled data, we propose a novel dual-task-consistency strategy, which learns from unlabeled data by minimizing the difference between the predicted pixel-wise label and the level set function. To build the consistency, a transform layer is used to convert the level set function to a pixel-wise probability map, which is implemented by smooth Heaviside function. In the following two subsections, we first introduce the dual-task consistency strategy, then introduce the semi-supervised training for  segmentation through dual-task consistency.
\subsubsection{Dual-task Consistency:}In general semi-supervised learning, consistency losses are designed to encourage smooth predictions in a data-level, i.e. the predictions of same data under different transformations~\cite{li2020transformation} and perturbations~\cite{ouali2020semi} should be the same. In contrast to data-level consistency, we enforce the task-level consistency between the pixel-level classification task, defined as task1 and the level set regression task, defined as task2. In existing works, pixel-wise classification for segmentation has been widely studied while level-set function \citep{li2005level} is a traditional task that captures geometric active contours and distance information, which rejuvenates recently when combining with CNN \citep{wang2020deep}. We introduce the level set function defined as follows:
\floatname{Algorithm 1}{Semi-supervised training through Dual-task consistency}  
\renewcommand{\algorithmicrequire}{\textbf{Input: }}  
\renewcommand{\algorithmicensure}{\textbf{Output:}}  
    \begin{algorithm}[tb]  
  \caption{Semi-supervised training through Dual-task consistency,}  
%   \LinesNumbered
  \begin{algorithmic}[1]  
    \Require  
      $\mathbf{x}_{i} \in \mathcal{D}_{l} + \mathcal{D}_{n}$,
      $\mathbf{y}_{i} \in \mathcal{D}_{l}$
     
    \Ensure  
      Dual-task model's parameter $\theta_{1}$ for segmentation head, $\theta_{2}$ for level-set function (LSF) head and $\theta$ for shared-weights backbone network
    %\State $f_{1}\left(x\right)$ is the segmentation task.
    %\State $f_{2}\left(x\right)$ is the LSF task.
    \State $f_{1}\left(x\right)$ = segmentation task branch with shared parameter $\theta$ and segmentation head's parameter $\theta_{1}$
    \State $f_{2}\left(x\right)$ = LSF task branch with shared parameter $\theta$ and LSF head's parameter $\theta_{2}$
    \While{ stopping criterion not met:}
    \State     Sample batch $b_l = (\mathbf{x}_{i},\mathbf{y}_{i}) \in \mathcal{D}_{l}$ and $b=b_l+b_u$, where $b_u = \mathbf{x}_{i} \in \mathcal{D}_{u}$
    \State Generating LSF ground truth $\mathcal{T}(\mathbf{y}_{i})$ according to Equation.~\ref{equ:level-set-func}
    \State Computing dual-task predictions $f_{1}\left(\mathbf{x}_{i}\right)$ and $f_{2}\left(\mathbf{x}_{i}\right)$, $i \in \{1, ..., N\}$ where $N$ denotes the batch size
    \State Applying task transform layer  $\mathcal{T}^{-1} \left(f_{2}(\mathbf{x}_{i})\right)$ according to Equation.~\ref{equ:trans-func}
    \State $\mathcal{L}_{DTC}(\mathbf{x}) =\frac{1}{\left|b\right|}\sum_{\mathbf{x}_{i} \in b }\left\|f_{1}(\mathbf{x}_{i})-\mathcal{T}^{-1}\left(f_{2}(\mathbf{x}_{i})\right)\right\|^{2} $
    \State $\mathcal{L}_{LSF}(\mathbf{x}, \mathbf{y})=\frac{1}{\left|b_{l}\right|}\sum_{\mathbf{x}_{i},\mathbf{y}_{i} \in b_l}\left\|f_{2}(\mathbf{x}_{i})- \mathcal{T}(\mathbf{y}_{i})\right\|^{2}$
    \State $\mathcal{L}_{Seg}(\mathbf{x}, \mathbf{y})
=1-\frac{1}{\left|b_{l}\right|}\sum_{\mathbf{x}_{i}, \mathbf{y}_{i} \in b_{l}} 2 \frac{\sum f_{1}\left(\mathbf{x}_{i}\right)\mathbf{y}_{i}}{\sum f_{1}\left(\mathbf{x}_{i}\right) + \sum \mathbf{y}_{i}}$

    \label{code:fram:add}  
    \State $\mathcal{L}_{total}=\mathcal{L}_{Seg} + \mathcal{L}_{LSF} + \lambda_d\mathcal{L}_{DTC}$
     
    \State Computing gradient of loss function $\mathcal{L}_{total}$ and update network parameters $\theta_{1}$, $\theta_{2}$ and $\theta$ by back propagation.
    \EndWhile \\
    \Return  $\theta_{1}$, $\theta_{2}$ and $\theta$ 
  \end{algorithmic} 
 \label{alg:Framwork}  
\end{algorithm}  

\begin{equation}
\mathcal{T}(x) =\left\{\begin{aligned}\label{equ:level-set-func}
& -\inf _{y \in \partial S}\|x-y\|_{2}, & x \in \mathcal{S}_{\mathrm{in}} 
\\& 0, & x \in \mathcal{\partial S} 
\\& +\inf _{y \in \partial S}\|x-y\|_{2}, & x \in \mathcal{S}_{\mathrm{out}}
\end{aligned}\right.
\end{equation}where $x$, $y$ are two different pixels/voxels in a segmentation mask, the $\partial S$ is the zero level set and also represents the contour of the target object. $\mathcal{S}_{\mathrm{in}}$ and $\mathcal{S}_{\mathrm{out}}$ denote the inside region and outside region of the target object. Then we define $\mathcal{T}(x)$ as the task transform from segmentation map to level-set function map in Equation.~\ref{equ:level-set-func}. To map the output of LSF task to the space of segmentation output, it is natural to think of using an inverse transform of $\mathcal{T}(x)$. However, it is impractical to integrate the exact inverse transform of $\mathcal{T}(x)$ in training due to the
non-differentiability. Hence, we utilize a smooth approximation to the inverse transform of level-set function, provided that we want to guarantee the values of $\mathcal{S}_{\mathrm{in}}$ are assigned to 1 while those of $\mathcal{S}_{\mathrm{out}}$ are assigned to 0 in the transformed prediction map, which is defined as:
\begin{equation}\label{equ:trans-func}
\mathcal{T}^{-1}(z)=\frac{1}{1+e^{-k \cdot z}}=\sigma(k \cdot z)
\end{equation}where $z$ means the level set value at pixel/voxel $x$. The formulation of $\mathcal{T}^{-1}(z)$ is delicate and simple as it is equal to Sigmoid function with the input multiplied by a factor $k$, which is selected as large as possible to approximate inverse transform of $\mathcal{T}(x)$. Thus, $\mathcal{T}^{-1}(z)$ can easily be implemented as an modified activate function followed by task2's output. Then the differentiability can be proved as follows:
\begin{equation}\begin{aligned}
\frac{\partial \mathcal{T}^{-1}}{\mathrm{d} z}&=\left(\frac{1}{1+e^{-k \cdot z}}\right)^{\prime} \\
&=k \cdot  \frac{1}{1+e^{-kz}} \cdot\left(1-\frac{1}{1+e^{-kz}}\right) \\
\end{aligned}
\end{equation}Though such approximate transform function will map the prediction space of task2 to be the same with that of task1, it naturally introduces a task-level prediction difference since task1 focuses on pixel-level reasoning while task2 attends to geometric structure information.
 Thus, for input $\mathbf{X}$ from a dataset $\mathcal{D}$, we define the dual-task-consistency loss $\mathcal{L}_{DTC}$ enforcing consistency between task1's prediction $f_{1}(\mathbf{x}_{i})$ and the transformed map of task2's prediction $\mathcal{T}^{-1}\left(f_{2}(\mathbf{x}_{i})\right)$ :

  \begin{equation}
  \begin{aligned}
\mathcal{L}_{DTC}(\mathbf{x}) &=\sum_{\mathbf{x}_{i} \in \mathcal{D} }\left\|f_{1}(\mathbf{x}_{i})-\mathcal{T}^{-1}\left(f_{2}(\mathbf{x}_{i})\right)\right\|^{2} \\
&=\sum_{\mathbf{x}_{i} \in \mathcal{D}}\left\|f_{1}(\mathbf{x}_{i})- \sigma\left(k \cdot f_{2}(\mathbf{x}_{i})\right)\right\|^{2}
\end{aligned}
 \end{equation}
%  Figure.~\ref{fig:method} demonstrates how to construct the dual-task consistency via task transform layer. To show the pixel-wise response, we illustrate the predicted, transformed and GT probability maps for one 2D slice from a 3D left artrial volume, when the training process is almost done.
\subsubsection{Semi-supervised training through Dual-Task-Consistency:}

\begin{figure*}
    \centering
    \includegraphics[width=1.0\textwidth]{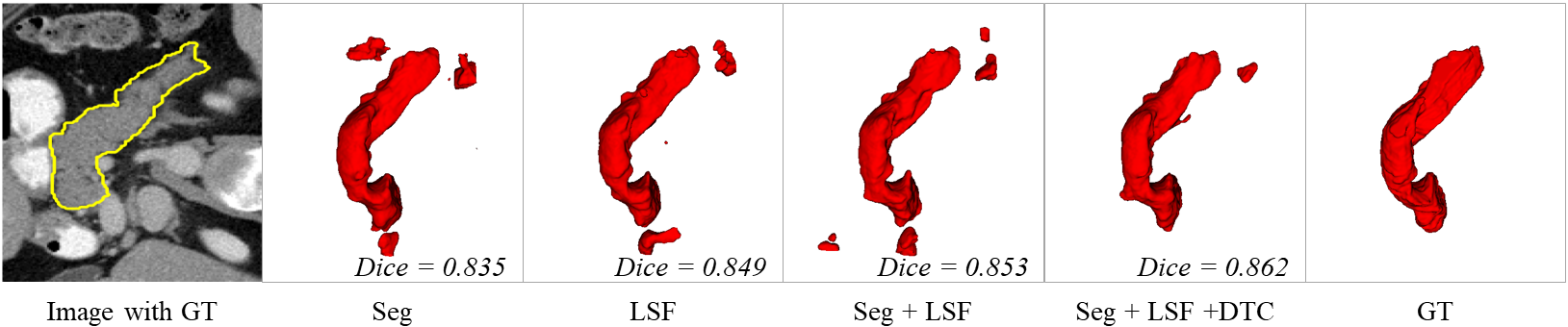}
    \caption{3D Visualization of different training methods for pancreas segmentation. 12 annotated images without unannotated images were used for training. GT: ground truth. (best viewed in color)}
    \label{fig:abla-visualization}
\end{figure*}

\begin{table*}[htb]
\centering
\footnotesize
\setlength{\tabcolsep}{1.3mm}{
\begin{tabular}{c|c|c|c|c|c|c|c|c
}
\hline
\multirow{2}{*}{Method} & \multicolumn{2}{c|}{Scans used} & \multicolumn{4}{c|}{Metrics} & \multicolumn{2}{c}{Cost}    \\ \cline{2-9} 
& Labeled & Unlabeled & Dice (\%) & Jaccard (\%) &ASD (voxel) & 95HD (voxel) & Params (M) & Training time (h) \\ \hline
Seg &12 & 0& 70.63  &56.72  & 6.29 & 22.54 & 9.44 & \textbf{2.1}\\ 

LSF &12&0&71.78 & 57.55 & 6.31 &20.74 &9.44 & 2.1 \\ 

Seg + LSF& 12& 0& 73.08& 58.65& 4.47&18.04 &9.44 & 2.2\\ 

Seg + LSF + DTC &12 &0 &\textbf{74.84} & \textbf{60.78}& \textbf{2.17} & \textbf{9.34} & 9.44 & 2.3\\ 
\hline
Seg &62 & 0& 81.78  &69.65  & 1.34 & 5.13 & 9.44 & \textbf{2.3}\\ 
LSF &62&0& 82.25&70.23 &\textbf{1.18} &5.19 &9.44 & 2.5 \\ 
Seg + LSF& 62& 0& 82.46& 70.61& 1.22&4.97 &9.44 &  2.5\\ 
Seg + LSF + DTC &62 &0 &\textbf{82.80} & \textbf{71.05}& 1.45 & \textbf{4.67} & 9.44 & 2.5\\ 

\hline
\end{tabular}}
\caption{Ablation study of our dual-task consistency method on the Pancreas CT dataset.}\label{tab:abla_tab}
\end{table*}

\begin{figure*}[tbp]
    \centering
    \includegraphics[width=1.0\textwidth]{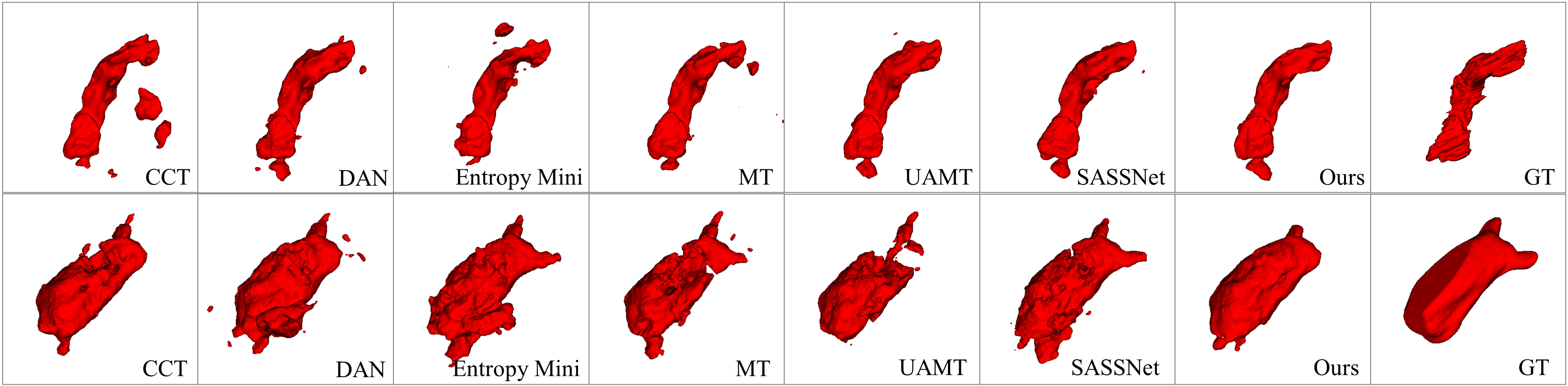}
    \caption{3D Visualization of different semi-supervised segmentation methods under 20\% labeled data  (best viewed in color). The first row is a pancreas segmentation result and second row is a left atrium segmentation result.}
    \label{fig:sota-visualization}
\end{figure*}

Let $\mathcal{D}_{l}$ and $\mathcal{D}_{u}$ be the labeled and unlabeled dataset, respectively. Let 
$\mathcal{D} = \mathcal{D}_{l} \cup \mathcal{D}_{u}$ be the whole provided dataset.
We denote labeled data pair as $(\mathbf{X}, \mathbf{Y}) \in \mathcal{D}_{l}$ and unlabeled data as $\mathbf{X} \in \mathcal{D}_{u}$, where $\mathbf{Y}$ is groundtruth segmentation mask. We denote voxel-level pair as $(x, y) \in (\mathbf{X}, \mathbf{Y})$. For labeled data $\mathcal{D}_{l}$, we define the supervised loss for segmentation task as commonly used dice loss :
\begin{equation}
\begin{aligned}
&\mathcal{L}_{Seg}(\mathbf{x}, \mathbf{y})
= \sum_{\mathbf{x}_{i}, \mathbf{y}_{i} \in \mathcal{D}_{l}} \mathcal{L}_{Dice}(\mathbf{{x}_{i}}, \mathbf{y}_{i})\\
&=\sum_{\mathbf{x}_{i}, \mathbf{y}_{i} \in \mathcal{D}_{l}} (1- \frac{2 \sum_{x_{j} \in \mathbf{x}_{i}, y_{j} \in \mathbf{y}_{i}} f_{1}\left(x_{i}\right)y_{i}}{\sum_{x_{j} \in \mathbf{x}_{i}, y_{j} \in \mathbf{y}_{i}} f_{1}\left(x_{j}\right) + \sum_{y_{j} \in \mathbf{y}_{i}} y_{j}})
\end{aligned}
\end{equation}
where the summation for $\sum_{x_{j} \in \mathbf{x}_{i}, y_{j} \in \mathbf{y}_{i}}$ denotes voxel-wise sum in a 3D image, and the summation for $\sum_{\mathbf{x}_{i}, \mathbf{y}_{i} \in \mathcal{D}_{l}}$ denotes image-level sum in a dataset. Then we define the supervised loss for LSF task as $\mathcal{L}_{2}$ loss between the predicted probability map $f_{2}(\mathbf{x})$ and the transformed ground truth map $\mathcal{T}(\mathbf{y})$:
\begin{equation}
\mathcal{L}_{LSF}(\mathbf{x}, \mathbf{y})=\sum_{\mathbf{x}_{i},\mathbf{y}_{i} \in \mathcal{D}_l}\left\|f_{2}(\mathbf{x}_{i})- \mathcal{T}(\mathbf{y}_{i})\right\|^{2}
\end{equation}
It is noteworthy that for annotated images, the ground truth level set function for the LSF task can be automatically generated from labeled segmentation mask $\mathbf{Y}$ through aforementioned task transform function $\mathcal{T}$. The final loss is defined as:
\begin{equation}\label{equ:total_loss}
\mathcal{L}_{total}=\mathcal{L}_{Seg} + \mathcal{L}_{LSF} + \lambda_d\mathcal{L}_{DTC}
\end{equation}
where $\mathcal{L}_{Seg}$ and $\mathcal{L}_{LSF}$ are only used for labeled data, while $\mathcal{L}_{DTC}$ is used for both labeled and unlabeled data during training, and therefore the two tasks can jointly optimize the network with either labeled data or unlabeled data in a semi-supervised fashion. Following~\citep{tarvainen2017mean,yu2019uncertainty}, we use a time-dependent Gaussian warming up function $\lambda_d(t) = e^{(-5(1-\frac{t}{t_{max}})^2)}$ to control the balance between the supervised loss and unsupervised consistency loss, where $t$ denotes the current training step and $t_{max}$ is the maximum training step. The used training algorithm for semi-supervised segmentation through dual-task consistency is shown in Algorithm.~\ref{alg:Framwork}.

\section{Experiments and Results}
\textbf{Datasets and Pre-processing: }To evaluate the proposed method, we apply our algorithm on two different datasets. The first is left atrial dataset~\cite{xiong2020global}, which consists of 100 3D gadolinium-enhanced MR images, with a resolution of $0.625\times0.625\times0.625 mm$. Following ~\citep{yu2019uncertainty,li2020shape}, we use 80 scans for training and 20 scans for validation, and apply the same pre-processing methods. The second is pancreas dataset~\cite{roth2015deeporgan}, which includes 82 abdomen CT images. Following~\citep{xia20203d}, we randomly split them into 62 images for training and 20 images for testing. In pre-processing, we use the soft tissue CT window range of [$-125$, $275$] HU~\cite{zhou2019prior}, and resample all images to an isotropic resolution of $1.0\times1.0\times1.0 mm$. Finally, we crop the images centering at the pancreas
region based on the ground truth with enlarged margins (25 voxels) and normalize them as
zero mean and unit variance. In this work, we report the performance of all methods trained with 20\% labeled images and 80\% unlabeled images, which is the typical semi-supervised learning experimental setting~\citep{xia20203d,yu2019uncertainty,li2020shape}.

\begin{figure}[tbp]
    \centering
    \includegraphics[width=0.46\textwidth]{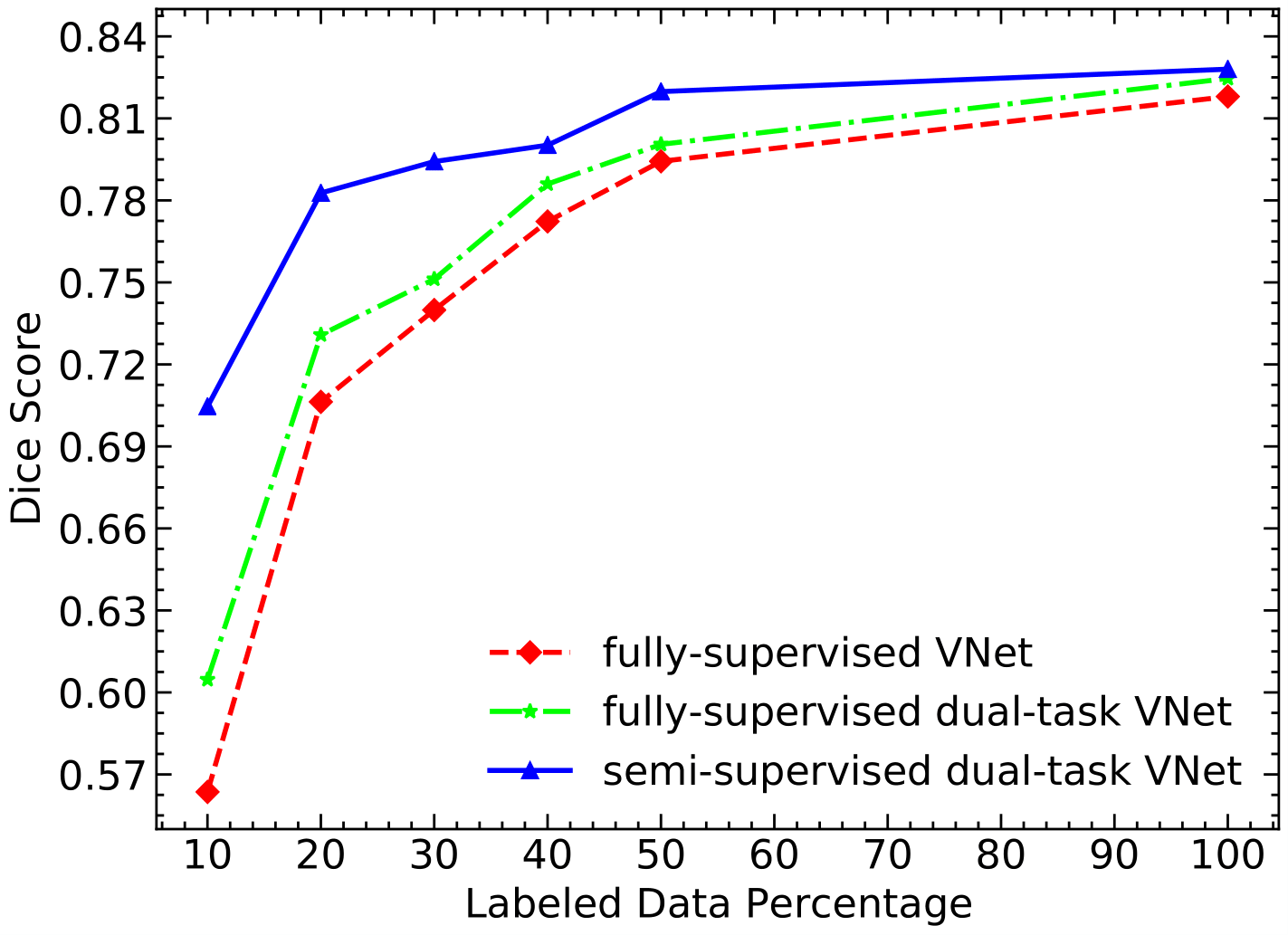}
    \caption{The pancreas segmentation performance of our semi-supervised approach with different ratio of labeled data. The dashed red and lime curves show performance of fully-supervised VNet and dual-task VNet respectively, where they were trained with only the available labeled data.}
    \label{fig:plot-ratio}
\end{figure}

\subsubsection{Implementation Details and Evaluation Metrics: }We implement our framework in PyTorch~\citep{paszke2019pytorch}, using an NVIDIA 1080TI GPU. In this work, we use VNet~\citep{milletari2016v} as the backbone for all experiments, and we implement dual-task VNet by adding a new regression layer at the end of the original VNet. The framework is trained by an SGD optimizer for 6000 iterations, with an initial learning rate (lr) 0.01
decayed by 0.1 every 2500 iterations. The batch size is 4, consisting of 2 labeled images and 2 unlabeled images. Following~\citep{Xue2020ShapeAwareOS}, the value of $k$ is set to 1500 in this work. We randomly crop $112\times112\times80$ (3D MRI Left Atrium) and $96\times96\times96$ (3D CT Pancreas) sub-volume as the network input. To avoid over-fitting, we use the standard on-the-fly
data augmentation methods during training stage~\citep{yu2019uncertainty}. Note that, in this work, the level set function is generated before the training phase rather on-the-fly, since the level set function is transform-invariant, which in result significantly speed up the training procedure. In the inference phase, we use a sliding window strategy to obtain the final results, which with a stride of $18\times18\times4$ for left atrium and $16\times16\times16$ for pancreas. At the inference time, we use the output of pixel-wise classification branch as the segmentation result. For a fair comparison, we do not use any post-processing or ensemble methods. Following~\citep{yu2019uncertainty}, we use four metrics to quantitatively evaluate our method, including Dice, Jaccard, the average surface
distance (ASD), and the 95\% Hausdorff Distance (95HD).

\subsubsection{The Effects of Different Tasks: }To investigate the individual impact of different tasks, we first only use labeled images for training and analyze how the dual-task consistency performs when only labeled images are used. We trained the network for pancreas segmentation using the 12 labeled data and all the 62 labeled data, respectively. We compared different training strategies: 1) only using the branch for task1 (Seg), 2) only using the branch for task 2 (LSF), 3) using the two branches for task1 and task2 simultaneously (Seg + LSF), and 4) and our proposed dual-task consistency method (Seg + LSF + DTC). The performance of these variants is listed in Table.~\ref{tab:abla_tab}. It shows that the level set function regression is helpful for medical image segmentation. It also can be observed that dual-task consistency consistently improves the performance of the dual-task VNet on 12 labeled scans and 62 labeled scans. Figure.~\ref{fig:abla-visualization} shows some visualization of different training methods, which
further show the superiority of our proposed dual-task-consistency.

\begin{table*}[htbp]
\centering
\footnotesize
\renewcommand\arraystretch{1}%{}
\setlength{\tabcolsep}{1.3mm}{
\begin{tabular}{c|c|c|c|c|c|c|c|c
}
\hline
\multirow{2}{*}{Method} & \multicolumn{2}{c|}{Scans used} & \multicolumn{4}{c|}{Metrics} & \multicolumn{2}{c}{Cost}    \\ \cline{2-9} 
& Labeled & Unlabeled & Dice (\%) & Jaccard (\%) &ASD (voxel) & 95HD (voxel) & Params (M) & Training time (h) \\ \hline
VNet & 12 & 0 &70.63  &56.72  & 6.29 & 22.54 & 9.44 & \textbf{2.1}\\
VNet & 62 & 0 &81.78  &69.65  & 1.34 & 5.13 & 9.44 &  2.3\\ 

\hline
MT (NeurIPS'17) & 12 & 50 & 75.85 & 61.98
 & 3.40 & 12.59  & 9.44 & 2.9\\ 
DAN (MICCAI'17) & 12 & 50 & 76.74 & 63.29 & 2.97& 11.13 & 12.09  & 3.3\\ 
Entropy Mini (CVPR'19) & 12 & 50 &75.31  &61.73  & 3.88 & 11.72 & 9.44 & \textbf{2.2}\\ 
UA-MT (MICCAI'19) & 12 & 50 & 77.26 & 63.82 & 3.06
 & 11.90 & 9.44 & 3.9\\ 
CCT (CVPR'20) & 12 & 50 & 76.58 & 62.76 & 3.69 & 12.92 & 15.65 & 4.1\\

SASSNet (MICCAI'20) & 12 & 50 & 77.66 & 64.08 & 3.05 & 10.93 & 20.46 & 3.9\\
\hline
Ours & 12 & 50 & \textbf{78.27} & \textbf{64.75} & \textbf{2.25} & \textbf{8.36} & \textbf{9.44} & 2.5\\

\hline
\end{tabular}}
\caption{Quantitative comparison between our methods and other semi-supervised methods on the Pancreas CT dataset. The first and second row are our fully supervised baseline, the last row is our proposed method, others are previous methods.}
\label{tab:panc-sota}
\end{table*}

\begin{table*}[tbp]
\centering
\footnotesize
\renewcommand\arraystretch{1}%{}
\setlength{\tabcolsep}{1.3mm}{
\begin{tabular}{c|c|c|c|c|c|c|c|c
}
\hline
\multirow{2}{*}{Method} & \multicolumn{2}{c|}{Scans used} & \multicolumn{4}{c|}{Metrics} & \multicolumn{2}{c}{Cost}    \\ \cline{2-9} 
% & Labeled & Unlabeled & Dice (\%) & Jaccard (\%)　& ASD (voxel) & 95HD (voxel) & Params (M) & Training time (h) \\ \hline
& Labeled & Unlabeled & Dice (\%) & Jaccard (\%) &ASD (voxel) & 95HD (voxel) & Params (M) & Training time (h) \\ \hline
VNet & 16 & 0 & 86.03 & 73.26 & 5.75 & 17.93 & 9.44 & \textbf{1.8}\\ 
VNet & 80 & 0 & 91.14 & 83.32 & 1.52 & 5.75 & 9.44 & 2.0\\ 

\hline
MT(NeurIPS'17) & 16 & 64 & 88.23 & 79.29 & 2.73 &  10.64 & 9.44 &3.2\\ 
DAN (MICCAI'17) & 16 & 64 &87.52 & 78.29& 2.42 & 9.01& 12.09 & 3.7\\ 

Entropy Mini (CVPR'19) & 16 & 64 & 88.45 & 79.51 & 3.72 & 14.14 & 9.44 & \textbf{1.9}\\

UA-MT (MICCAI'19) & 16 & 64 & 88.88 & 80.21 & 2.26 & 7.32 & 9.44 &3.6\\ 

CCT (CVPR'20) & 16 & 64 & 88.83 & 80.06 & 2.49 & 8.44 &15.65 & 3.9\\
SASSNet (MICCAI'20) & 16 & 64 & 89.27& 80.82& 3.13 & 8.83 & 20.46 & 4.4\\
\hline
Ours & 16 & 64 & \textbf{89.42} & \textbf{80.98} & \textbf{2.10} & \textbf{7.32} & \textbf{9.44} & 2.2\\

\hline
\end{tabular}}
\caption{Quantitative comparison between our methods and other semi-supervised methods on the Left Atrium MRI dataset. The first and second row are our fully supervised baseline, the last row is our proposed method, others are previous methods.}
\label{tab:la-sota}
\end{table*}

\subsubsection{Effectiveness of Dual-task-Consistency for Semi-supervised Learning: }Secondly, we performed a study on data utilization efficiency of our approach compared to the fully supervised VNet and dual-task VNet that only use available annotated images for training on Pancreas CT dataset. We draw the Dice score of the results in Figure.\ref{fig:plot-ratio}. It can be observed that the semi-supervised method consistently performs better than the supervised approach in different labeled data settings, demonstrating that our method effectively utilizes the unlabeled data and brings performance gains. It also can be  found that the performance gap between fully supervised method and semi-supervised approach narrows as more labeled images are available, which conforms to the common sense. When the number of labeled data is small, our method also can obtain a better segmentation result than fully supervised method, indicating the promising potential of our proposed approach for further clinical use.

\subsubsection{Comparison with Other Semi-supervised Methods:} We compared our framework with six state-of-the-art semi-supervised segmentation methods, including deep adversarial network (DAN)~\citep{zhang2017deep}, entropy minimization approach (Entropy Mini)~\citep{vu2019advent}, cross-consistency training method (CCT)~\citep{ouali2020semi}, mean teacher self-ensembling model (MT)~\citep{tarvainen2017mean}, uncertainty-aware mean teacher model(UA-MT)~\citep{yu2019uncertainty} and shape-aware adversarial network (SASSNet)~\citep{li2020shape}. Note that we used the official code and results of DAN, MT, UA-MT, SASSNet, and reimplemented the Entropy Mini and CCT for medical image segmentation, since the limitation of GPU memory, we used one main decoder and three auxiliary decoders as CCT's implementation. 

\par We first evaluate our proposed framework on Pancreas CT. Table.~\ref{tab:panc-sota} shows the quantitative comparison of these methods. Compared with fully supervised VNet trained with only 12 annotated images, all semi-supervised methods taking advantages of unannotated images improve the segmentation performance significantly. The MT, UA-MT and CCT achieve slightly better performance than Entropy Mini and DAN, demonstrating that perturbation-based consistency loss is helpful
for the semi-supervised segmentation problem. Moreover, the UA-MT is better than MT, since the uncertainty map can guide the student model learning efficiently. The SASSNet achieves the top performance among the existing methods, indicating the shape prior is useful for semi-supervised image segmentation. Notably, our framework achieves better performance than the state-of-the-art semi-supervised methods on all the evaluation metrics without using a complex multiple network architecture, corroborating that our dual-task-consistency has the full capability to draw out the rich information from the unlabeled data. Meanwhile, our framework does not require any multiple inference or iteratively update scheme, which reduces the computational memory cost and running time.
\par We further validate our proposed method on Left Atrium MRI data, which is a widely-used dataset for semi-supervised medical image segmentation~\citep{yu2019uncertainty,li2020shape}. A quantitative comparison of these methods is shown in Tabel.~\ref{tab:la-sota}. It can be found that our method achieved the best accuracy than other methods on all the evaluation metrics, especially in term of ASD and 95HD. Figure.~\ref{fig:sota-visualization} shows some visualization of pancreas segmentation and left atrium segmentation. Compared with other methods, our results have higher overlap ratio with the ground truth and produce less false positives and preserve more details, which further indicates the effectiveness, generalization and robustness of our proposed method. Furthermore, we investigated the training cost of different approaches. The quantitative comparison of network's parameters and training time are listed in Table.\ref{tab:panc-sota} and Table.\ref{tab:la-sota}. It can be observed that, our framework requires less training time than MT, DAN, UAMT, CCT and SASSNet, since our framework use a simple network with fewer parameters and does not need to pass an image many times in an iteration. Compared with Entropy Mini and fully supervised baseline, our method achieved better accuracy with comparable computational cost. Thus, our experiments prove that our method attains the best accuracy, networks' parameters and computational-cost trade-offs.
 
\section{Discussion and Conclusion}
In this paper, we have presented a novel and simple semi-supervised medical image segmentation framework through dual-task consistency, which is a task-level consistency-based framework for semi-supervised segmentation. We use a dual-task network that simultaneously predicts a pixel-level classification map and a level set representation of the segmentation that is able to capture global-level shape and geometric information. In order to build a semi-supervised training framework, we enforce dual-task consistency between classification map prediction and LSF prediction via a task-transform layer. We achieve stat-of-the-art results on two 3D medical image datasets including left atrial dataset in MR scans and pancreas dataset in CT scans. The superior performance demonstrates the effectiveness, robustness and generalization of our proposed framework. In this work, we focus
on single-class segmentation to simplify the presentation.
However, our method extends to the multi-class case in
a straightforward manner.

In addition, our proposed method can easily be extended to use additional tasks such as edge extraction~\cite{zhen2020joint} and key-points estimation~\cite{cheng2020higherhrnet} as long as there exists differentiable transform between two tasks. 
We also hope to inspire the whole computer vision community, as it is possible to construct tasks consistency in a semi-supervised fashion in many directions such as two-stream video recognition~\cite{simonyan2014two}, multi-task image reconstruction~\cite{zamir2018taskonomy,zamir2020robust} etc. to leverage a large amount of unlabeled data. In the future, we will extend this method to more computer vision applications to reduce labeling efforts and further investigate the fusion strategy to ensemble all different tasks' prediction results for better performance.
\section{Acknowledgments} This work was supported by the National Natural Science Foundations of China [81771921, 61901084], and also by key research and
development project of Sichuan province, China [20ZDYF2817]. We would like to thank Mr. Yechong Huang for constructive discussions, suggestion and manuscript proofread and also thank the organization teams of MICCAI 2018 left atrial segmentation challenge, the National Institutes of Health Clinical Center for the publicly available datasets.
\bibliography{ref}

\begin{thebibliography}{44}
\providecommand{\natexlab}[1]{#1}
\providecommand{\url}[1]{\texttt{#1}}
\providecommand{\urlprefix}{URL }
\expandafter\ifx\csname urlstyle\endcsname\relax
  \providecommand{\doi}[1]{doi:\discretionary{}{}{}#1}\else
  \providecommand{\doi}{doi:\discretionary{}{}{}\begingroup
  \urlstyle{rm}\Url}\fi

\bibitem[{Bai et~al.(2017)Bai, Oktay, Sinclair, Suzuki, Rajchl, Tarroni,
  Glocker, King, Matthews, and Rueckert}]{bai2017semi}
Bai, W.; Oktay, O.; Sinclair, M.; Suzuki, H.; Rajchl, M.; Tarroni, G.; Glocker,
  B.; King, A.; Matthews, P.~M.; and Rueckert, D. 2017.
\newblock Semi-supervised learning for network-based cardiac MR image
  segmentation.
\newblock In \emph{MICCAI}, 253--260. Springer.

\bibitem[{Chen et~al.(2018)Chen, Zhu, Papandreou, Schroff, and
  Adam}]{chen2018encoder}
Chen, L.-C.; Zhu, Y.; Papandreou, G.; Schroff, F.; and Adam, H. 2018.
\newblock Encoder-decoder with atrous separable convolution for semantic image
  segmentation.
\newblock In \emph{ECCV}, 801--818.

\bibitem[{Cheng et~al.(2020)Cheng, Xiao, Wang, Shi, Huang, and
  Zhang}]{cheng2020higherhrnet}
Cheng, B.; Xiao, B.; Wang, J.; Shi, H.; Huang, T.~S.; and Zhang, L. 2020.
\newblock HigherHRNet: Scale-Aware Representation Learning for Bottom-Up Human
  Pose Estimation.
\newblock In \emph{CVPR}, 5386--5395.

\bibitem[{Dai, He, and Sun(2015)}]{dai2015boxsup}
Dai, J.; He, K.; and Sun, J. 2015.
\newblock Boxsup: Exploiting bounding boxes to supervise convolutional networks
  for semantic segmentation.
\newblock In \emph{CVPR}, 1635--1643.

\bibitem[{Jeong et~al.(2019)Jeong, Lee, Kim, and Kwak}]{jeong2019consistency}
Jeong, J.; Lee, S.; Kim, J.; and Kwak, N. 2019.
\newblock Consistency-based semi-supervised learning for object detection.
\newblock In \emph{NeurIPS}, 10759--10768.

\bibitem[{Kr{\"a}henb{\"u}hl and Koltun(2011)}]{krahenbuhl2011efficient}
Kr{\"a}henb{\"u}hl, P.; and Koltun, V. 2011.
\newblock Efficient inference in fully connected crfs with gaussian edge
  potentials.
\newblock In \emph{NeurIPS}, 109--117.

\bibitem[{Lee et~al.(2019)Lee, Kim, Lee, Lee, and Yoon}]{lee2019ficklenet}
Lee, J.; Kim, E.; Lee, S.; Lee, J.; and Yoon, S. 2019.
\newblock Ficklenet: Weakly and semi-supervised semantic image segmentation
  using stochastic inference.
\newblock In \emph{CVPR}, 5267--5276.

\bibitem[{Li et~al.(2005)Li, Xu, Gui, and Fox}]{li2005level}
Li, C.; Xu, C.; Gui, C.; and Fox, M.~D. 2005.
\newblock Level set evolution without re-initialization: a new variational
  formulation.
\newblock In \emph{CVPR}, volume~1, 430--436. IEEE.

\bibitem[{Li, Zhang, and He(2020)}]{li2020shape}
Li, S.; Zhang, C.; and He, X. 2020.
\newblock Shape-aware Semi-supervised 3D Semantic Segmentation for Medical
  Images.
\newblock In \emph{MICCAI}, 552--561. Springer.

\bibitem[{Li et~al.(2020)Li, Yu, Chen, Fu, Xing, and
  Heng}]{li2020transformation}
Li, X.; Yu, L.; Chen, H.; Fu, C.-W.; Xing, L.; and Heng, P.-A. 2020.
\newblock Transformation-Consistent Self-Ensembling Model for Semisupervised
  Medical Image Segmentation.
\newblock \emph{IEEE Transactions on Neural Networks and Learning Systems} .

\bibitem[{Lin et~al.(2016)Lin, Dai, Jia, He, and Sun}]{lin2016scribblesup}
Lin, D.; Dai, J.; Jia, J.; He, K.; and Sun, J. 2016.
\newblock Scribblesup: Scribble-supervised convolutional networks for semantic
  segmentation.
\newblock In \emph{CVPR}, 3159--3167.

\bibitem[{Long, Shelhamer, and Darrell(2015)}]{long2015fully}
Long, J.; Shelhamer, E.; and Darrell, T. 2015.
\newblock Fully convolutional networks for semantic segmentation.
\newblock In \emph{CVPR}, 3431--3440.

\bibitem[{Ma, He, and Yang(2020)}]{ma2020TMI-LGAC}
Ma, J.; He, J.; and Yang, X. 2020.
\newblock Learning Geodesic Active Contours for Embedding Object Global
  Information in Segmentation CNNs.
\newblock \emph{IEEE Transactions on Medical Imaging} .

\bibitem[{Ma et~al.(2020)Ma, Wei, Zhang, Wang, Lv, Zhu, Chen, Liu, Peng, Wang
  et~al.}]{ma2020distance}
Ma, J.; Wei, Z.; Zhang, Y.; Wang, Y.; Lv, R.; Zhu, C.; Chen, G.; Liu, J.; Peng,
  C.; Wang, L.; et~al. 2020.
\newblock How Distance Transform Maps Boost Segmentation CNNs: An Empirical
  Study.
\newblock In \emph{MIDL}.

\bibitem[{Masood et~al.(2019)Masood, Fang, Li, Li, Sheng, Mathavan, Wang, Yang,
  Wu, Qin et~al.}]{masood2019automatic}
Masood, S.; Fang, R.; Li, P.; Li, H.; Sheng, B.; Mathavan, A.; Wang, X.; Yang,
  P.; Wu, Q.; Qin, J.; et~al. 2019.
\newblock Automatic choroid layer segmentation from optical coherence
  tomography images using deep learning.
\newblock \emph{Scientific reports} 9(1): 1--18.

\bibitem[{Masood et~al.(2015)Masood, Sharif, Masood, Yasmin, and
  Raza}]{masood2015survey}
Masood, S.; Sharif, M.; Masood, A.; Yasmin, M.; and Raza, M. 2015.
\newblock A survey on medical image segmentation.
\newblock \emph{Current Medical Imaging Reviews} 11(1): 3--14.

\bibitem[{Milletari, Navab, and Ahmadi(2016)}]{milletari2016v}
Milletari, F.; Navab, N.; and Ahmadi, S.-A. 2016.
\newblock V-net: Fully convolutional neural networks for volumetric medical
  image segmentation.
\newblock In \emph{3DV}, 565--571. IEEE.

\bibitem[{Nie et~al.(2018)Nie, Gao, Wang, and Shen}]{nie2018asdnet}
Nie, D.; Gao, Y.; Wang, L.; and Shen, D. 2018.
\newblock ASDNet: Attention based semi-supervised deep networks for medical
  image segmentation.
\newblock In \emph{MICCAI}, 370--378. Springer.

\bibitem[{Ouali, Hudelot, and Tami(2020)}]{ouali2020semi}
Ouali, Y.; Hudelot, C.; and Tami, M. 2020.
\newblock Semi-Supervised Semantic Segmentation with Cross-Consistency
  Training.
\newblock In \emph{CVPR}, 12674--12684.

\bibitem[{Paszke et~al.(2019)Paszke, Gross, Massa, Lerer, Bradbury, Chanan,
  Killeen, Lin, Gimelshein, Antiga et~al.}]{paszke2019pytorch}
Paszke, A.; Gross, S.; Massa, F.; Lerer, A.; Bradbury, J.; Chanan, G.; Killeen,
  T.; Lin, Z.; Gimelshein, N.; Antiga, L.; et~al. 2019.
\newblock Pytorch: An imperative style, high-performance deep learning library.
\newblock In \emph{NeurIPS}, 8026--8037.

\bibitem[{Portela, Cavalcanti, and Ren(2014)}]{portela2014semi}
Portela, N.~M.; Cavalcanti, G.~D.; and Ren, T.~I. 2014.
\newblock Semi-supervised clustering for MR brain image segmentation.
\newblock \emph{Expert Systems with Applications} 41(4): 1492--1497.

\bibitem[{Qiao et~al.(2018)Qiao, Shen, Zhang, Wang, and Yuille}]{qiao2018deep}
Qiao, S.; Shen, W.; Zhang, Z.; Wang, B.; and Yuille, A. 2018.
\newblock Deep co-training for semi-supervised image recognition.
\newblock In \emph{ECCV}, 135--152.

\bibitem[{Ronneberger, Fischer, and Brox(2015)}]{ronneberger2015u}
Ronneberger, O.; Fischer, P.; and Brox, T. 2015.
\newblock U-net: Convolutional networks for biomedical image segmentation.
\newblock In \emph{MICCAI}, 234--241. Springer.

\bibitem[{Roth et~al.(2015)Roth, Lu, Farag, Shin, Liu, Turkbey, and
  Summers}]{roth2015deeporgan}
Roth, H.~R.; Lu, L.; Farag, A.; Shin, H.-C.; Liu, J.; Turkbey, E.~B.; and
  Summers, R.~M. 2015.
\newblock Deeporgan: Multi-level deep convolutional networks for automated
  pancreas segmentation.
\newblock In \emph{MICCAI}, 556--564. Springer.

\bibitem[{Sajjadi, Javanmardi, and Tasdizen(2016)}]{sajjadi2016regularization}
Sajjadi, M.; Javanmardi, M.; and Tasdizen, T. 2016.
\newblock Regularization with stochastic transformations and perturbations for
  deep semi-supervised learning.
\newblock In \emph{NeurIPS}, 1163--1171.

\bibitem[{Simonyan and Zisserman(2014)}]{simonyan2014two}
Simonyan, K.; and Zisserman, A. 2014.
\newblock Two-stream convolutional networks for action recognition in videos.
\newblock In \emph{NeurIPS}, 568--576.

\bibitem[{Song et~al.(2020)Song, Chen, Luo, Huang, Liu, Huang, Chen, Ye, Sheng,
  Zhang et~al.}]{song2020cpm}
Song, T.; Chen, J.; Luo, X.; Huang, Y.; Liu, X.; Huang, N.; Chen, Y.; Ye, Z.;
  Sheng, H.; Zhang, S.; et~al. 2020.
\newblock CPM-Net: A 3D Center-Points Matching Network for Pulmonary Nodule
  Detection in CT Scans.
\newblock In \emph{International Conference on Medical Image Computing and
  Computer-Assisted Intervention}, 550--559. Springer.

\bibitem[{Tarvainen and Valpola(2017)}]{tarvainen2017mean}
Tarvainen, A.; and Valpola, H. 2017.
\newblock Mean teachers are better role models: Weight-averaged consistency
  targets improve semi-supervised deep learning results.
\newblock In \emph{NeurIPS}, 1195--1204.

\bibitem[{Vu et~al.(2019)Vu, Jain, Bucher, Cord, and P{\'e}rez}]{vu2019advent}
Vu, T.-H.; Jain, H.; Bucher, M.; Cord, M.; and P{\'e}rez, P. 2019.
\newblock Advent: Adversarial entropy minimization for domain adaptation in
  semantic segmentation.
\newblock In \emph{CVPR}, 2517--2526.

\bibitem[{Wang et~al.(2018{\natexlab{a}})Wang, Li, Zuluaga, Pratt, Patel,
  Aertsen, Doel, David, Deprest, Ourselin et~al.}]{bifseg}
Wang, G.; Li, W.; Zuluaga, M.~A.; Pratt, R.; Patel, P.~A.; Aertsen, M.; Doel,
  T.; David, A.~L.; Deprest, J.; Ourselin, S.; et~al. 2018{\natexlab{a}}.
\newblock Interactive medical image segmentation using deep learning with
  image-specific fine tuning.
\newblock \emph{IEEE Transactions on Medical Imaging} 37(7): 1562--1573.

\bibitem[{Wang et~al.(2018{\natexlab{b}})Wang, Zuluaga, Li, Pratt, Patel,
  Aertsen, Doel, David, Deprest, Ourselin et~al.}]{deepigeos}
Wang, G.; Zuluaga, M.~A.; Li, W.; Pratt, R.; Patel, P.~A.; Aertsen, M.; Doel,
  T.; David, A.~L.; Deprest, J.; Ourselin, S.; et~al. 2018{\natexlab{b}}.
\newblock DeepIGeoS: a deep interactive geodesic framework for medical image
  segmentation.
\newblock \emph{IEEE Transactions on Pattern Analysis and Machine Intelligence}
  41(7): 1559--1572.

\bibitem[{Wang et~al.(2020)Wang, Wei, Liu, Chen, Zhou, Shen, Fishman, and
  Yuille}]{wang2020deep}
Wang, Y.; Wei, X.; Liu, F.; Chen, J.; Zhou, Y.; Shen, W.; Fishman, E.~K.; and
  Yuille, A.~L. 2020.
\newblock Deep distance transform for tubular structure segmentation in ct
  scans.
\newblock In \emph{CVPR}, 3833--3842.

\bibitem[{Xia et~al.(2020)Xia, Liu, Yang, Cai, Yu, Zhu, Xu, Yuille, and
  Roth}]{xia20203d}
Xia, Y.; Liu, F.; Yang, D.; Cai, J.; Yu, L.; Zhu, Z.; Xu, D.; Yuille, A.; and
  Roth, H. 2020.
\newblock 3d semi-supervised learning with uncertainty-aware multi-view
  co-training.
\newblock In \emph{WACV}, 3646--3655.

\bibitem[{Xiong et~al.(2020)Xiong, Xia, Hu, Huang, Vesal, Ravikumar, Maier, Li,
  Tong, Si et~al.}]{xiong2020global}
Xiong, Z.; Xia, Q.; Hu, Z.; Huang, N.; Vesal, S.; Ravikumar, N.; Maier, A.; Li,
  C.; Tong, Q.; Si, W.; et~al. 2020.
\newblock A Global Benchmark of Algorithms for Segmenting Late
  Gadolinium-Enhanced Cardiac Magnetic Resonance Imaging.
\newblock \emph{Medical Image Analysis} .

\bibitem[{Xue et~al.(2020)Xue, Tang, Qiao, Gong, Yin, Qian, Huang, Fan, and
  Huang}]{Xue2020ShapeAwareOS}
Xue, Y.; Tang, H.; Qiao, Z.; Gong, G.; Yin, Y.; Qian, Z.; Huang, C.; Fan, W.;
  and Huang, X. 2020.
\newblock Shape-Aware Organ Segmentation by Predicting Signed Distance Maps.
\newblock In \emph{AAAI}.

\bibitem[{You et~al.(2011)You, Peng, Yuan, Cheung, and
  Lei}]{you2011segmentation}
You, X.; Peng, Q.; Yuan, Y.; Cheung, Y.-m.; and Lei, J. 2011.
\newblock Segmentation of retinal blood vessels using the radial projection and
  semi-supervised approach.
\newblock \emph{Pattern recognition} 44(10-11): 2314--2324.

\bibitem[{Yu et~al.(2019)Yu, Wang, Li, Fu, and Heng}]{yu2019uncertainty}
Yu, L.; Wang, S.; Li, X.; Fu, C.-W.; and Heng, P.-A. 2019.
\newblock Uncertainty-aware self-ensembling model for semi-supervised 3D left
  atrium segmentation.
\newblock In \emph{MICCAI}, 605--613. Springer.

\bibitem[{Zamir et~al.(2020)Zamir, Sax, Cheerla, Suri, Cao, Malik, and
  Guibas}]{zamir2020robust}
Zamir, A.~R.; Sax, A.; Cheerla, N.; Suri, R.; Cao, Z.; Malik, J.; and Guibas,
  L.~J. 2020.
\newblock Robust Learning Through Cross-Task Consistency.
\newblock In \emph{CVPR}, 11197--11206.

\bibitem[{Zamir et~al.(2018)Zamir, Sax, Shen, Guibas, Malik, and
  Savarese}]{zamir2018taskonomy}
Zamir, A.~R.; Sax, A.; Shen, W.; Guibas, L.~J.; Malik, J.; and Savarese, S.
  2018.
\newblock Taskonomy: Disentangling task transfer learning.
\newblock In \emph{CVPR}, 3712--3722.

\bibitem[{Zhang et~al.(2017)Zhang, Yang, Chen, Fredericksen, Hughes, and
  Chen}]{zhang2017deep}
Zhang, Y.; Yang, L.; Chen, J.; Fredericksen, M.; Hughes, D.~P.; and Chen, D.~Z.
  2017.
\newblock Deep adversarial networks for biomedical image segmentation utilizing
  unannotated images.
\newblock In \emph{MICCAI}, 408--416. Springer.

\bibitem[{Zhen et~al.(2020)Zhen, Wang, Zhou, Li, Shen, Shang, Fang, and
  Quan}]{zhen2020joint}
Zhen, M.; Wang, J.; Zhou, L.; Li, S.; Shen, T.; Shang, J.; Fang, T.; and Quan,
  L. 2020.
\newblock Joint Semantic Segmentation and Boundary Detection using Iterative
  Pyramid Contexts.
\newblock In \emph{CVPR}, 13666--13675.

\bibitem[{Zhou et~al.(2019{\natexlab{a}})Zhou, Li, Bai, Wang, Chen, Han,
  Fishman, and Yuille}]{zhou2019prior}
Zhou, Y.; Li, Z.; Bai, S.; Wang, C.; Chen, X.; Han, M.; Fishman, E.; and
  Yuille, A.~L. 2019{\natexlab{a}}.
\newblock Prior-aware neural network for partially-supervised multi-organ
  segmentation.
\newblock In \emph{ICCV}, 10672--10681.

\bibitem[{Zhou et~al.(2019{\natexlab{b}})Zhou, Wang, Tang, Bai, Shen, Fishman,
  and Yuille}]{zhou2019semi}
Zhou, Y.; Wang, Y.; Tang, P.; Bai, S.; Shen, W.; Fishman, E.; and Yuille, A.
  2019{\natexlab{b}}.
\newblock Semi-supervised 3D abdominal multi-organ segmentation via deep
  multi-planar co-training.
\newblock In \emph{WACV}, 121--140. IEEE.

\bibitem[{Zhu et~al.(2020)Zhu, Li, Hu, Ma, Zhou, and Zheng}]{zhu2020rubik}
Zhu, J.; Li, Y.; Hu, Y.; Ma, K.; Zhou, S.~K.; and Zheng, Y. 2020.
\newblock Rubik’s Cube+: A Self-supervised Feature Learning Framework for 3D
  Medical Image Analysis.
\newblock \emph{Medical Image Analysis} 101746.

\end{thebibliography}
\end{document}